\documentclass{mfa}

\usepackage{mathtools}
\usepackage{amsmath}
\usepackage{amsfonts}
\usepackage{latexsym}
\usepackage{color}
\usepackage{graphicx}
\usepackage{subcaption}
\usepackage{units}
\usepackage{listings}
\usepackage{courier}
\usepackage{verbatim}
\usepackage{bigints}
\graphicspath{ {/home/gabe/schoolstuff/BDA} }

\DeclarePairedDelimiterX\Basics[1](){ #1}

\usepackage{scalerel,stackengine}

\usepackage{xspace}
\newcommand{\vx}{\mathbf{x}}
\newcommand{\rx}{\mathrm{x}}
\newcommand{\mX}{\mathbf{X}}

\stackMath
\newcommand\reallywidehat[1]{\savestack{\tmpbox}{\stretchto{  \scaleto{    \scalerel*[\widthof{\ensuremath{#1}}]{\kern-.6pt\bigwedge\kern-.6pt}    {\rule[-\textheight/2]{1ex}{\textheight}}  }{\textheight}}{0.5ex}}\stackon[1pt]{#1}{\tmpbox}}
\begin{document}

\title{What do Asian Religions Have in Common?\\
An Unsupervised Text Analytics Exploration }
\author[Sah]{Preeti Sah}
\address{College of Computing and Information Sciences, Rochester Institute of Technology, 85 Lomb Memorial Drive, Rochester, New York 14623}
\email{ks3911@rit.edu}

\author[Fokou\'e \\]{Ernest Fokou\'e \\}
\address{School of Mathematical Sciences, Rochester Institute of Technology, 85 Lomb Memorial Drive, Rochester, New York 14623}
\email{epfeqa@rit.edu}

{62F07}

\setlength\parindent{0pt}
\setlength{\overfullrule}{0pt}

\newtheorem*{theorem}{Theorem}
\newtheorem*{lemma}{Lemma}
\newtheorem*{example}{Example}
\newtheorem*{assumption}{Assumption}
\newtheorem*{proposition}{Proposition}

\begin{abstract}
The main source of various religious teachings is their sacred texts which varies from religion to religion based on different factors like the geographical region or time of birth of particular religion. Despites these differences there could be similarities between the sacred texts based on what lessons it teaches to it's followers. This paper attempts to find the similarity using text mining techniques. The corpus consisting of Asian (Tao Te Ching, Buddhism, Yogasutra, Upanishad) and non Asian (four Bible texts) is used to explore findings of similarity measures like Euclidean, Manhattan, Jaccard and Cosine on raw Document Term Frequency [DTM], normalized DTM which reveals similarity based on word usage. The performance of Supervised learning algorithms like K-Nearest Neighbor [KNN], Support Vector Machine [SVM] and Random Forest is measured based on it's accuracy to predict correct scared text for any given chapter in the corpus. The K-means clustering visualizations on Euclidean distances of raw DTM reveals that there exists a pattern of similarity among these sacred texts with Upanishads and Tao Te Ching being the most similar text in the corpus.
\end{abstract}

\maketitle

\section{Introduction}
The purpose of religion is to facilitate love, compassion, patience, tolerance, humility and forgiveness. The sacred texts are cornerstone of religion and medium to instill the religious teachings in the people. Every part of the world follow different sacred texts to learn and preach about their religion.\newline

The following scripts were collected for different religions which is followed in different countries:
\begin{itemize} 
  \item Hinduism (India): Yogasutras, Upanishads 
  \item Buddhism (Tibet): Four Noble Truth of Buddhism 
  \item Taoism (China): Tao Te Ching
  \item Christianity (Central Asia/America): Book of Proverb, Book of Ecclesiastes, Book of Ecclesiasticus, Book of Wisdom
\end{itemize}

All the data collected was English translations of the original language in which it was written.This was done to make sure that we have uniformity of texts collected from different sources. 

The sources of the data were: 
\begin{itemize}
  \item Yogasutras: Project Gutenberg's The Yoga Sutras of Patanjali, by Charles Johnston
  \item Upanishads: The Project Gutenberg EBook of The Upanishads, by Swami Paramananda
  \item Four Noble Truth of Buddhism: https://www.accesstoinsight.org/lib/study/truths.html
  \item Tao Te Ching: Tao Te Ching - Translated by J. Legge
  \item Book of Proverb: Project Gutenberg EBook The Bible, Douay-Rheims, Book 22: Proverbs
  \item Book of Ecclesiastes: Project Gutenberg EBook The Bible, Douay-Rheims, Book 23: Ecclesiastes
  \item Book of Ecclesiasticus: Project Gutenberg EBook The Bible, Douay-Rheims, Book 26: Ecclesiasticus
  \item Book of Wisdom: Project Gutenberg EBook The Bible, Douay-Rheims, Book 25: Wisdom

\end{itemize}

Buddhism teaches about four noble truth. Each of these truths entails a duty: stress is to be comprehended, the origination of stress abandoned, the cessation of stress realized, and the path to the cessation of stress developed. When all of these duties have been fully performed, the mind gains total release \cite{Budd}. Tao Te Ching teaches that Tao is The Way, Not ‘Your Way’about. The chapters talk about staying detached, letting go and keeping things simple \cite{Tao}.
Yogasutra contains essence of wisdom. We think of ourselves as living a purely physical life, in these material bodies of ours. In reality, we have gone far indeed from pure physical
life; for ages, our life has been psychical, we have been centred and
immersed in the psychic nature \cite{Yoga}. The Upanishads represent the loftiest heights of ancient
Indo-Aryan thought and culture. They form the wisdom portion or
Gnana-Kanda of the Vedas, as contrasted with the Karma-Kanda or
sacrificial portion. In each of the four great Vedas--known as
Rik, Yajur, Sama and Atharva--there is a large portion which
deals predominantly with rituals and ceremonials, and which has
for its aim to show man how by the path of right action he may
prepare himself for higher attainment \cite{Upanishad}. \newline
Book of Proverbs consists of wise and weighty
sentences: regulating the morals of men: and directing them to wisdom
and virtue \cite{Proverb}. Book of Ecclesiastes or The Preacher, (in Hebrew,
Coheleth,) because in it, Solomon, as an excellent preacher, setteth
forth the vanity of the things of this world: to withdraw the hearts and
affections of men from such empty toys \cite{Ecclesiastes}. Book of Ecclesiasticus gives admirable lessons of all virtues \cite{Ecclesiasticus}. Book of Wisdom abounds with instructions and
exhortations to kings and all magistrates to minister justice in the
commonwealth, teaching all kinds of virtues under the general names of
justice and wisdom \cite{Wisdom}.\newline

Buddhism :{\it And what are fabrications? There are these six classes of intention: intention aimed at sights, sounds, aromas, tastes, tactile sensations,  ideas. These are called fabrications.}
Tao Te Ching:  {\it Heaven and earth do not act from (the impulse of) any wish to be   benevolent; they deal with all things as the dogs of grass are dealt   with.  The sages do not act from (any wish to be) benevolent; they   deal with the people as the dogs of grass are dealt with.     May not the space between heaven and earth be compared to a   bellows?       'Tis emptied, yet it loses not its power;     'Tis moved again, and sends forth air the more.      Much speech to swift exhaustion lead we see;      Your inner being guard, and keep it free \cite{Budd}. } 

Tao Te Ching:  {\it Heaven and earth do not act from (the impulse of) any wish to be   benevolent; they deal with all things as the dogs of grass are dealt   with.  The sages do not act from (any wish to be) benevolent; they   deal with the people as the dogs of grass are dealt with.     May not the space between heaven and earth be compared to a   bellows?       'Tis emptied, yet it loses not its power;     'Tis moved again, and sends forth air the more.      Much speech to swift exhaustion lead we see;      Your inner being guard, and keep it free \cite{Tao}.} 

Upanishad : {\it The Brahman once won a victory for the Devas.  Through that victory of the Brahman, the Devas became elated.  They thought, "This victory is ours.  This glory is ours."  Brahman here does not mean a personal Deity.  There is a Brahma, the first person of the Hindu Trinity; but Brahman is the Absolute, the One without a second, the essence of all.  There are different names and forms which represent certain personal aspects of Divinity, such as Brahma the Creator, Vishnu the Preserver and Siva the Transformer; but no one of these can fully represent the Whole.  Brahman is the vast ocean of being, on which rise numberless ripples and waves of manifestation.  From the smallest atomic form to a Deva or an angel, all spring from that limitless ocean of Brahman, the inexhaustible Source of life.  No manifested form of life can be independent of its source, just as no wave, however mighty, can be independent of the ocean.  Nothing moves without that Power.  He is the only Doer.  But the Devas thought: "This victory is ours, this glory is ours." \cite{Upanishad} }

Yogasutra :  {\it perception of the true nature of things.  When the object is not truly perceived, when the observation is inaccurate and faulty, thought or reasoning based on that mistaken perception is of necessity false and unsound \cite{Yoga}.} 

Book of Proverb : {\it Doth not wisdom cry aloud, and prudence put forth her voice?  8:2. Standing in the top of the highest places by the way, in the midst of the paths,  8:3. Beside the gates of the city, in the very doors she speaketh, saying:  8:4. O ye men, to you I call, and my voice is to the sons of men.  8:5. O little ones understand subtlety, and ye unwise, take notice.  8:6. Hear, for I will speak of great things: and my lips shall be opened to preach right things.  8:7. My mouth shall meditate truth, and my lips shall hate wickedness \cite{Proverb}.}

Book of Ecclesiastes : {\it Speak not any thing rashly, and let not thy heart be hasty to utter a word before God. For God is in heaven, and thou upon earth: therefore let thy words be few.  5:2. Dreams follow many cares: and in many words shall be found folly.  5:3. If thou hast vowed any thing to God, defer not to pay it: for an unfaithful and foolish promise displeaseth him: but whatsoever thou hast vowed, pay it.  5:4. And it is much better not to vow, than after a vow not to perform the things promised.  5:5. Give not thy mouth to cause thy flesh to sin: and say not before the angel: There is no providence: lest God be angry at thy words, and destroy all the works of thy hands.  5:6. Where there are many dreams, there are many vanities, and words without number: but do thou fear God \cite{Ecclesiastes}.} 

Book of Ecclesiasticus : {\it Then Nathan the prophet arose in the days of David.  47:2. And as the fat taken away from the flesh, so was David chosen from among the children of Israel.  47:3. He played with lions as with lambs: and with bears he did in like manner as with the lambs of the flock, in his youth.  47:4. Did not he kill the giant, and take away reproach from his people?  47:5. In lifting up his hand, with the stone in the sling he beat down the boasting of Goliath:  47:6. For he called upon the Lord the Almighty, and he gave strength in his right hand, to take away the mighty warrior, and to set up the horn of his nation.  47:7. So in ten thousand did he glorify him, and praised him in the blessings of the Lord, in offering to him a crown of glory:  47:8. For he destroyed the enemies on every side, and extirpated the Philistines the adversaries unto this day: he broke their horn for ever.  47:9. In all his works he gave thanks to the holy one, and to the most High, with words of glory.  47:10. With his whole heart he praised the Lord, and loved God that made him: and he gave him power against his enemies:  47:11. And he set singers before the altar, and by their voices he made sweet melody \cite{Ecclesiasticus}.} 

Book of Wisdom : {\it Love justice, you that are the judges of the earth. Think of the Lord in goodness, and seek him in simplicity of heart:  1:2. For he is found by them that tempt him not: and he sheweth himself to them that have faith in him.  1:3. For perverse thoughts separate from God: and his power, when it is tried, reproveth the unwise:  1:4. For wisdom will not enter into a malicious soul, nor dwell in a body subject to sins.  1:5. For the Holy Spirit of discipline will flee from the deceitful, and will withdraw himself from thoughts that are without understanding, and he shall not abide when iniquity cometh in.  1:6. For the spirit of wisdom is benevolent, and will not acquit the evil speaker from his lips: for God is witness of his reins, and he is a true searcher of his heart, and a hearer of his tongue \cite{Wisdom}.} 

These texts from sacred scripts originated in different geographical locations and at different historic time-line.  The question arises is there are any similarity between them in terms what these texts want to teach and how they are teaching various religious lessons.\newline 
Text Mining using machine learning and feature extraction is helpful in finding patterns of words in document collections \cite{Qahl}. Using text mining the aim of this research is to find if various sacred texts are strongly connected. The similarity measures such as Euclidean, Manhattan, Jaccard and Cosine is firstly applied to word frequency matrix of the raw corpus to find similarities based on word usage. The distance matrices on Document Term Matrix formed by LDA was calculated to find the similarities between texts based on probabilistic models \cite{Cao} \cite{Romain} by selecting k topics \cite{Rajkumar}\cite{Thomas}. The unsupervised learning algorithm such as K mean clustering on raw frequency DTM reveals the strong similarity between sacred texts \cite{Bjornar}. Also supervised learning techniques like K-Nearest Neighbor, Support Vector Machine and Random Forest on labeled corpus was implemented to find if these algorithms can predict accurately if any chapter belongs to which sacred text.
\section{Methodology}
\label{motiv}
\subsection{Overview} 
\begin{figure}
\includegraphics[scale=0.5]{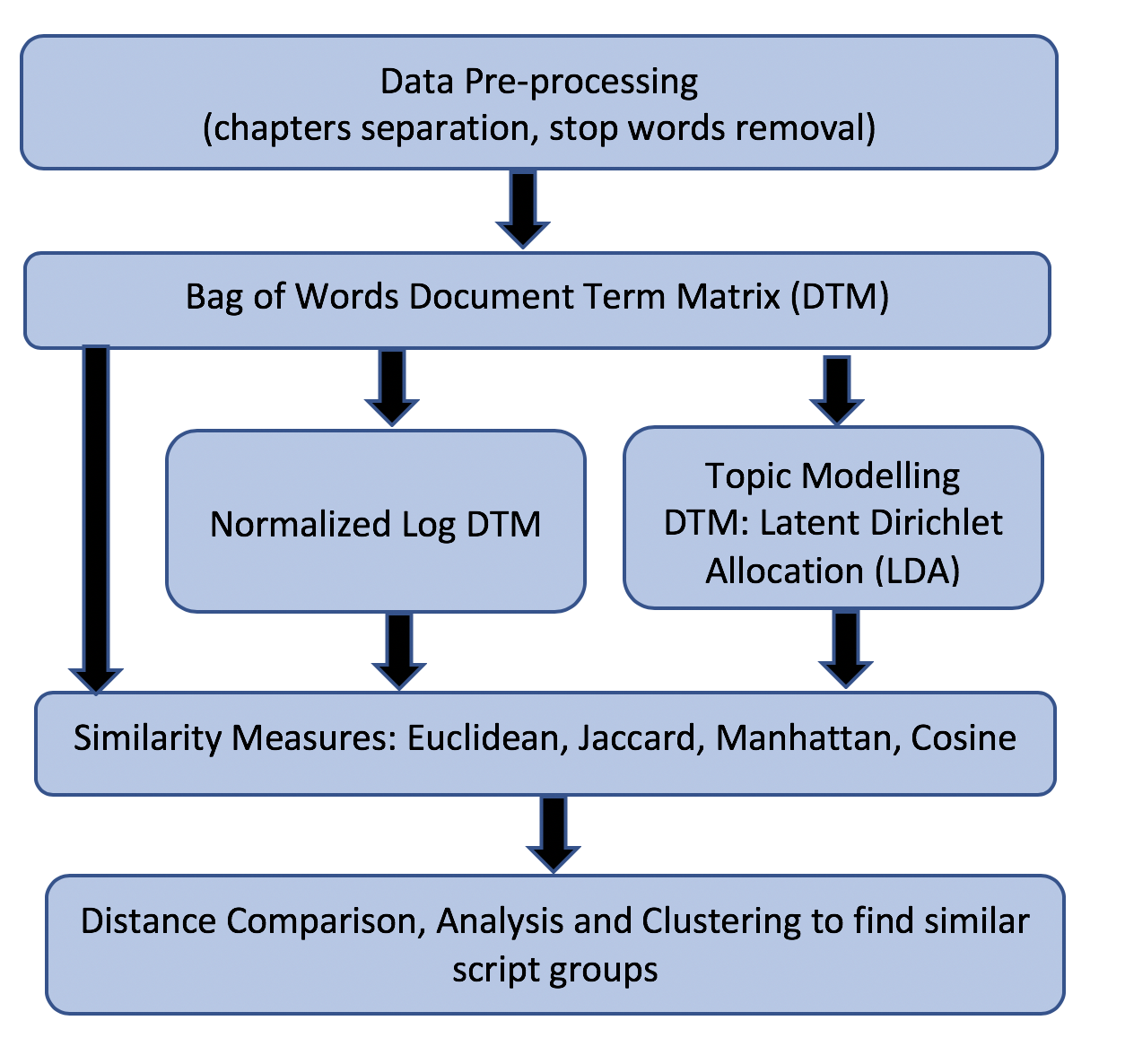}
 \centering
	\caption{Various steps involved in finding similarity between scriptures}
\label{fig:method}
\end{figure}

The Figure ~\ref{fig:method} shows overview of the steps and algorithms used to find similarity between religious scripts.

Bag of Words assumes that each document is the fragment of text from a sacred book. The distinction between sacred books is supervised via the creation of corresponding label. The closeness of sacred books is found in terms of document distances calculated using various similarity measures.

\subsection{Similarity Measures}
Throughout the rest of this paper, we will use the $p$-dimensional vector $\vx_l = (\rx_{l1}, \rx_{l2}, \cdots, \rx_{lp})^\top$ to denote the entries of the $l$th row of the term document matrix ${\bf X}$. Given two rows $\vx_l$ and $\vx_m$ of ${\bf X}$, we use the generic notation 
 $d(\vx_l, \vx_m)$ to denote the distance between the two rows, which is essentially the distance between two chosen chapters of the whole corpus regardless of which sacred book each belongs to. The chapter here is our basic document.
 
In this section we introduce different mathematical distances grouped mathematically
and we empirically evaluate their performance. Each distance family has specific mathematical properties that differentiates one another from each other. The effectiveness of
applying the similarity measure is believed to be related to the mathematical properties
of each family.\\

The various similarity measures helps us to understand the similarity of various chapters within the same book and also similarity between different book in the corpus.\\

The different measures used for the corpus:
\begin{itemize}
\item The very commonly known Euclidean distance belongs to Minkowski Family. The Euclidean distance between two chapters in the corpus is calculated as:

\begin{eqnarray}
d_E(\vx_l, \vx_m) = \left(\sum_{j=1}^p{(\rx_{lj}-\rx_{mj})^2}\right)^{\frac{1}{2}}
\end{eqnarray}

\item Manhattan distance belongs to Minkowski Family and distance between two chapters of the corpus is defined as: 

\begin{eqnarray}
d_M(\vx_l, \vx_m) = \sum_{j=1}^p{|\rx_{lj}-\rx_{mj}|}
\end{eqnarray}

\item Cosine Similarity measure is the normalized inner product between two documents on
the vector space that measures the cosine of the angle between them. The formula to find cosine similarity between two chapters can be written as:

\begin{eqnarray}
d_C(\vx_l, \vx_m) = \frac{\vx_l^\top\vx_m}{(\vx_l^\top\vx_l)^{\frac{1}{2}}(\vx_m^\top\vx_m)^{\frac{1}{2}}}
\end{eqnarray}

\item The Jaccard similarity measures the intersection between two chapters. Jaccard coefficient is calculated using the formula:

\begin{align*}
{\sf sim}(\vx_l,\vx_m)= \frac{\displaystyle \sum_{j=1}^{p}{\min\{\rx_{lj},\rx_{mj}\}}}{\displaystyle \sum_{k=1}^{p}{\max\{\rx_{lk},\rx_{mk}\}}}
\end{align*}

The Jaccard distance between two chapters is defined as:
\begin{align*}
d_J(\vx_l,\vx_m)= 1 - {\sf sim}(\vx_l,\vx_m)
\end{align*}
\end{itemize}

Using the above defined similarity measures on given books $X_a$ and $X_b$ we are trying to:

\begin{itemize}
	\item study $X_a$ or $X_b$ separately  \\
    
    $d(\mX_l^{(a)}, \mX_m^{(a)})\equiv$ distance between two chapters of same book $X^{(a)}$   
    
    This helps to discover the relationship of various chapters within the same book\\
    
    \item study relationship between $X_a$ or $X_b$ \\
  
\begin{eqnarray}  
d(\mX^{(a)}, \mX^{(b)}) = \underset{\substack{{\vx_l \in \mX^{(a)}}\\{\vx_m \in \mX^{(b)}}}}{\min}{\big\{d(\vx_l, \vx_m)\big\}}
\end{eqnarray}

    We are calculating the mean, median, minimum and maximum distances between chapters of different books to discover the relationship between the books. 
    \begin{equation*}
  d(X_a,X_b) =
    \begin{cases}
      \min \limits_{i \epsilon (1,..,n_a), j\epsilon (1,..,n_b)}\hspace{0.1 cm}{d(X_{ai},X_{bj})}\\
       
      \max \limits_{i \epsilon (1,..,n_a), j\epsilon (1,..,n_b)}\hspace{0.1 cm} {d(X_{ai},X_{bj})}\\
      \underset{i \epsilon (1,..,n_a), j\epsilon (1,..,n_b)}{\operatorname{average}} \hspace{0.1 cm}{d(X_{ai},X_{bj})}\\
      \underset{i \epsilon (1,..,n_a), j\epsilon (1,..,n_b)}{\operatorname{median}} \hspace{0.1 cm}{d(X_{ai},X_{bj})} \\
  
    \end{cases}    
\end{equation*}
   
  Within the book distance matrix helps to cluster the chapter in the same  book and is represented as: 

\[
D_X =
  \begin{bmatrix}
    d_{11} & d_{12} & d_{13} & \dots  & d_{1n} \\
    d_{21} & d_{22} & d_{23} & \dots  & d_{2n} \\
    \vdots & \vdots & \vdots & \ddots & \vdots \\
    d_{n1} & d_{n2} & d_{n3} & \dots  & d_{nn}
  \end{bmatrix}
\]
\hspace{2 cm}Distance between n chapters of script X \\

Distance matrix between eight books helps to cluster books across the corpus and is represented as:

\[
\Delta =
  \begin{bmatrix}
    X_{11} & X_{12} & X_{13} & \dots  & X_{18} \\
    X_{21} & X_{22} & X_{23} & \dots  & X_{28} \\
    \vdots & \vdots & \vdots & \ddots & \vdots \\
    X_{81} & X_{82} & X_{83} & \dots  & X_{88}
  \end{bmatrix}
\]

\end{itemize}

\subsection{Supervised Learning Algorithms}
Predictive aspects helps in prediction of the origin of fragments of spiritual literature. How well can we predict which sacred text a fragment of spiritual literature comes from? Three supervised algorithms: K-Nearest Neighbor, Support Vector Machine and Random Forest was applied on the labeled corpus. The supervised machines were trained on 70\% of the corpus and tested on remaining 30\%. The algorithm providing the maximum accuracy will be best in predicting the sacred text for a given chapter.

\subsection{Data Analysis}
Our goals with the data are 
\begin{itemize}
  \item Create a corpus where document is smallest unit of data
  \item Create Bag of Words DTM after data cleaning
  \item Attempt to confirm or discover some of the closeness among scared texts using similarity measures
  \item Measure the performance of supervised learning in identifying the book label for any document
\end{itemize}

There are several challenges with the data: non uniform structure data in each sacred book, initial preprocessing reveals large amount of stop words data which can mislead the similarity measures. Through this paper, document analysis assumes that (a) document is the smallest unit of data being used for finding similarity (b) within the bag of words (BOW) assumption/approach, each document is represented by the words. Using the BOW assumption, our basic data structure after pre-processing, is the term document matrix (tdm) also known as the document term matrix (dtm), which can be written in the following $n \times p$ matrix
\begin{equation}
		{\bf X} = \left[
		\begin{array}{ccccccccc}
		X_{11} & X_{12} & \cdots & \cdots & \cdots & \cdots & X_{1j} & \cdots & X_{1p}\\
		\vdots & \vdots & \ddots & \ddots & \cdots & \cdots & \cdots &\cdots & \vdots\\
		X_{i1} & X_{i2} & \cdots & \cdots & \cdots & \cdots & X_{ij} & \cdots & X_{ip}\\
		\vdots & \vdots & \ddots & \ddots & \cdots & \cdots & \cdots & \cdots & \vdots\\
		X_{n1} & X_{n2} & \cdots & \cdots & \cdots & \cdots & X_{nj} & \cdots & X_{np}\\
		\end{array}
		\right]
\label{eq:dtm:1}
		\end{equation}
        
Each column $X_j$ of ${\bf X}$ represents  a atomic word like {\it truth}, {\it diligent}, {\it sense}, {\it power}, {\it right}. In most document analysis tasks, the term document matrix ${\bf X}$ is typically very sparse, with $90\%$ of zeroes not unusual. Besides, except in rare cases, ${\bf X}$ tends to be ultra-high dimensional, meaning that $p \gg n$ as depicted in the matrix, since the number of words tends be much much higher than the number of documents to be text-analyzed.  Depending on the analysis, the entries $X_{ij}$ of ${\bf X}$ can be of one of the following types:
%
 \begin{itemize}
 \item $X_{ij}\equiv$ {\sf Frequency of word $j$ in document $i$.}
 \item $X_{ij}\equiv$ {\sf logarithmized relative frequency of word $j$ in document $i$.}
 \end{itemize}
As indicated earlier, one of the most interesting questions one may seek to answer in the presence of a collection of
documents dealing with the different sacred texts: {\it are there any similarity between the various sacred texts ? If so, can we measure that?} As we shall see later we will tackle this question using methods like {\it K-means clustering}. Specifically, if we anticipate $k$ groups of sacred texts, and denote by $P_k = C_1 \cup \cdots \cup C_k$, the partitioning of the data into $k$ groups/clusters, then we seek the optimum clustering.
\begin{eqnarray}
P_k^* = \underset{P_k}{\tt argmin}\left\{\sum_{j=1}^k{\sum_{i=1}^{n}{z_{ij}d(\vx_i,\vx_j^*)}}\right\},
\label{eq:clustering:1}
\end{eqnarray}
where $z_{ij}=\mathbb{L}(\vx_i \in C_j)$ and $d(\cdot)$ could be any distance like the Euclidean
$d(\vx_i,\vx_j^*)=\|\vx_i-\vx_j^*\|^2$ or the Manhattan distance $d(\vx_i,\vx_j^*)=\|\vx_i-\vx_j^*\|_1$, or any other suitable distance. Section $3$ of this paper is dedicated to the exploration of the clustering of the documents in our corpus. The other question that naturally arises from such a corpus of documents is: {\it For any given document can we predict which sacred text it belongs to?}  


\begin{itemize}

\item
{Data Processing}

The unstructured nature of text data adds an extra
layer of complexity in the feature extraction task, and the inherently sparse nature of
the corresponding data matrices makes text mining a distinctly difficult task. To deal with this problem it was required to process that data. There was a need to clean the noise using Natural Language processing (NLP). \linebreak

\item
{Data Cleaning}

The data cleaning involved removing of stop words using NLTK library. Apart from stop words present in library it was observed that the data required further cleaning. This was done by removing unnecessary punctuation marks, special characters and ancient English words which were not recognized as stop words by NLTK library.\linebreak

\item
{Data Sampling}

The organization of the text was: 
\begin{itemize}
	\item Books: Collection of entire script data
    \item Paragraphs: Division of script based on the topic being explained
    \item Chapters: Division of paragraph based on subtopic within each topic 
\end{itemize} 

Unit of Sampling: Chapter was taken as smallest unit of sampling. Each religious scripts was fragmented to chapters and stored for further process of finding the similarities.These units existed in the text such as Tao Te Ching while in other books it was approximated from texts headings. \\

Corpus $\equiv$ Various sacred texts 

Chapter $\equiv$  Collection of V words from corpus 

$Chapter_d$ $\equiv$ $x_d$ $\equiv$ ($x_{d1},x_{d2},..., x_{dv}$)

 \hspace{1 cm } $\equiv$ Input Vector 

\hspace{1 cm } $\equiv$ 1 chapter in a book \\

\item {Document Term Matrix (DTM) on Raw Text} 

The first input of similarity measures done using the raw corpus. The raw corpus in
this case refers to corpora after applying data cleaning and processing.
We are interested to handle the big corpus without any possible modification to test
distance measures performance. Hence, the term document matrix of the raw texts was
used as:

The rows of our term document matrix refer to a fragment of text from one of the sacred books, which is  a chapter in the sense adopted in this paper. The sacred book to which a document belongs is traced in a supervised manner with a variable $Y$ from the set of  labels of all the books considered here namely $\mathcal{Y} = \{g_1, g_2, \cdots, g_8\}$ where
\begin{itemize}
    \item []$g_1$ is Book $1$ containing chapters on {\sf the teachings of the Buddha}
    \item[]$g_2$ is Book 2 referring to the {\sf Tao Te Ching}
    \item[]$g_3$ is Book 3 referring to the {\sf Upanishads}
    \item[]$g_4$ is Book 4 referring to {\sf YogaSutra}
    \item[]$g_5$ is Book 5 referring to the {\sf Book of Proverb}
    \item[]$g_6$ is Book 6 referring to {\sf Book of Ecclesiastes}
    \item[]$g_7$ is Book 7 referring to {\sf Book of Ecclesiasticus}
    \item[]$g_8$ is Book 8 referring to {\sf Book of Wisdom}
\end{itemize}

\end{itemize}

\section{Results}
The minimum, maximum and average distances might contain outliers i.e chapters which are very similar to each other or quite dissimilar. To deal with this problem median of all distances of each chapter with every other chapter was used. The median distances was able to capture the  similarities which do not take outliers into consideration.\\

The Euclidean distance was able to separate the distances amongst different scripts while Cosine, Manhattan and Jaccard were unable to distinguish that.

\begin{figure}
\centering
        
\includegraphics{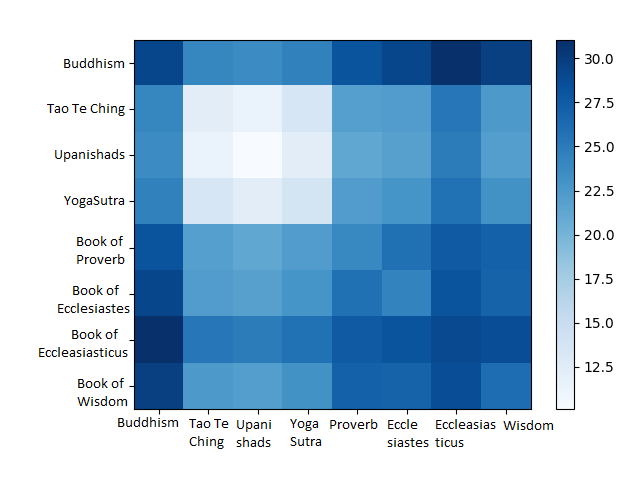}\caption{Euclidean median distance between different scripts}
\label{fig:eucldist}
\end{figure}

Figure~\ref{fig:eucldist} shows the Euclidean median distance of chapters within the same scripts and across the script. Between the scripts, distance is minimum between Upanishads and Tao Te Ching. Within the same script the distance of chapters within Upanishads is minimum (considering the diagonal).   

\clearpage
\pagebreak

\begin{figure}
    \centering
    \begin{subfigure}[b]{0.49\textwidth}
        \includegraphics[width=\textwidth]{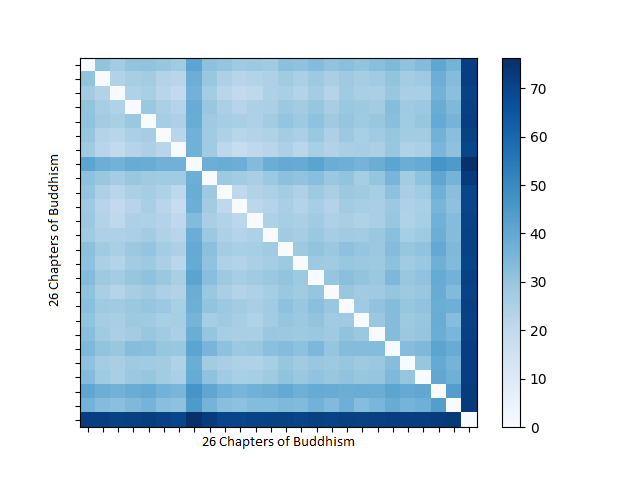}
        \caption{Buddhism}
        \label{fig:Buddhism}
    \end{subfigure}
    \begin{subfigure}[b]{0.49\textwidth}
        \includegraphics[width=\textwidth]{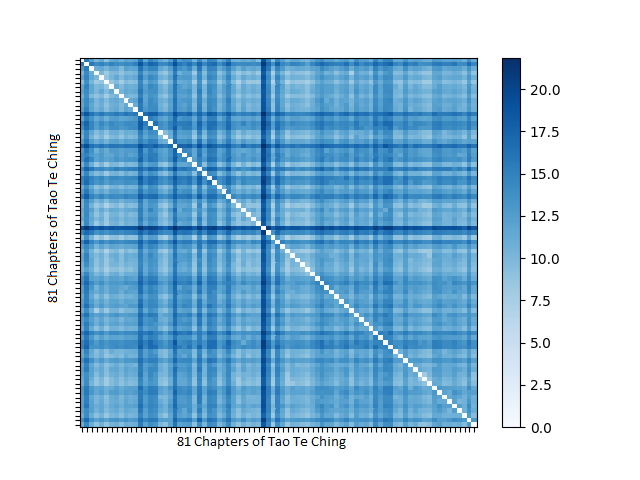}
        \caption{Tao Te Ching}
        \label{fig:Tao}
    \end{subfigure}
    
       \begin{subfigure}[b]{0.49\textwidth}
        \includegraphics[width=\textwidth]{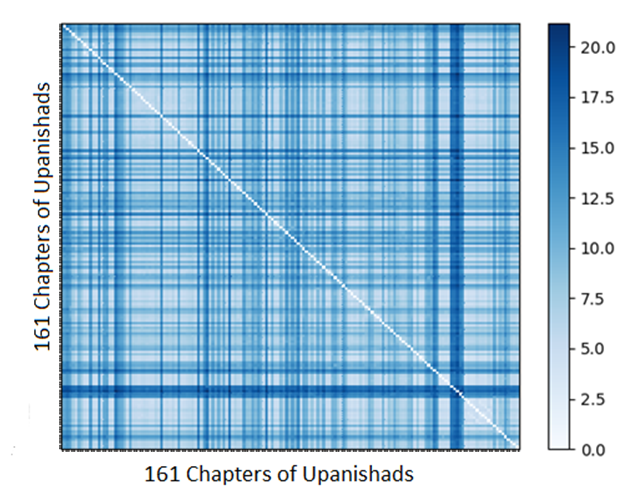}
        \caption{Upanishads}
        \label{fig:Upanishads}
    \end{subfigure}
    \begin{subfigure}[b]{0.49\textwidth}
        \includegraphics[width=\textwidth]{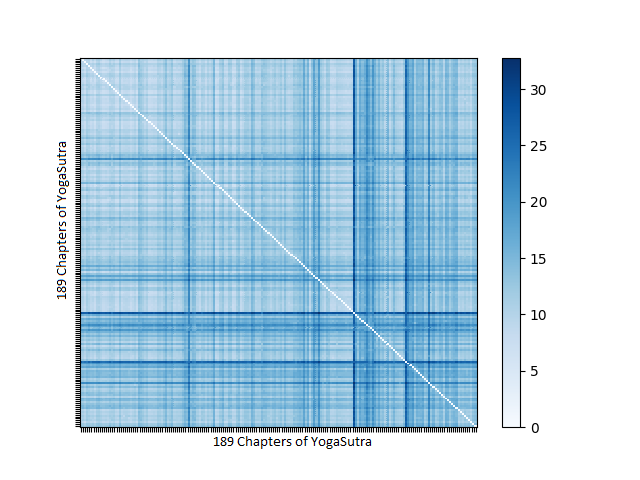}
        \caption{Yogasutra}
        \label{fig:Yogasutra}
    \end{subfigure}
    
    \caption{Euclidean distance between different chapters of Asian Religious scriptures}
    \label{fig:AsianScripts}
\end{figure}

Figure~\ref{fig:Buddhism}, ~\ref{fig:Tao},  ~\ref{fig:Upanishads} and ~\ref{fig:Yogasutra} the Euclidean distance of chapters within the Asian scriptures which helps to find most similar chapters within the same script.

\begin{figure}
    \centering
    \begin{subfigure}[b]{0.49\textwidth}
        \includegraphics[width=\textwidth]{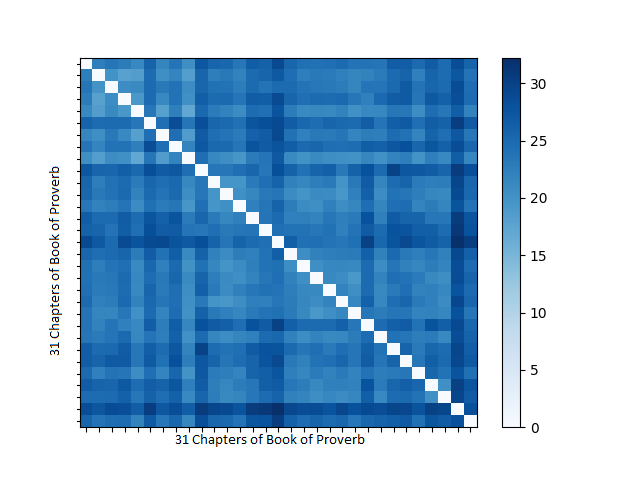}
        \caption{Book of Proverb}
        \label{fig:Proverb}
    \end{subfigure}
    \begin{subfigure}[b]{0.49\textwidth}
        \includegraphics[width=\textwidth]{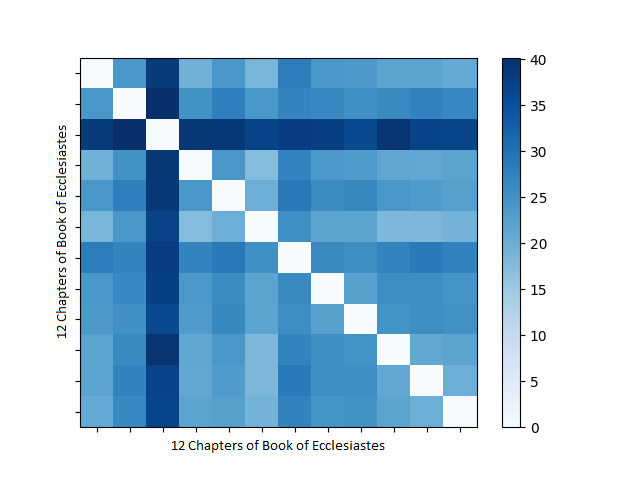}
        \caption{Book of Ecclesiastes}
        \label{fig:Ecc}
    \end{subfigure}
    
       \begin{subfigure}[b]{0.49\textwidth}
        \includegraphics[width=\textwidth]{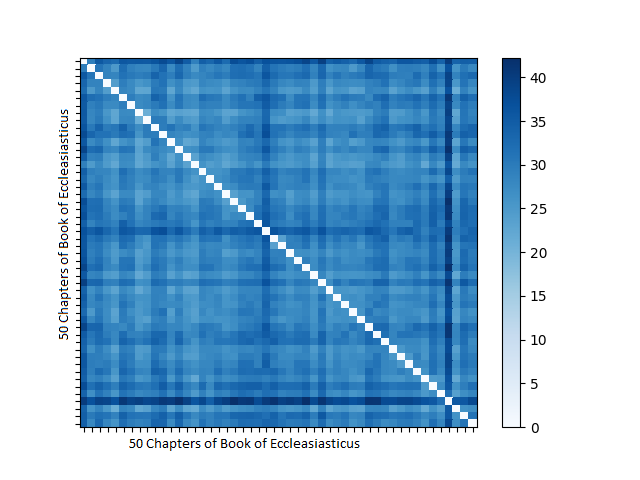}
        \caption{Book of Ecclesiasticus}
        \label{fig:Eccle}
    \end{subfigure}
    \begin{subfigure}[b]{0.49\textwidth}
        \includegraphics[width=\textwidth]{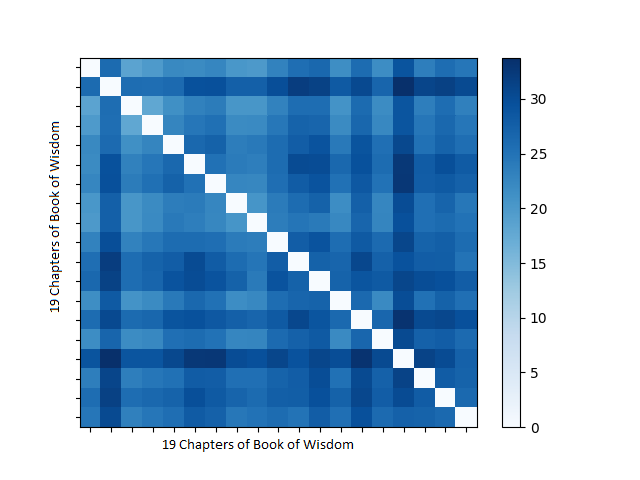}
        \caption{Book of Wisdom}
        \label{fig:wisdomchp}
    \end{subfigure}
    
    \caption{Euclidean distance between different chapters of Bible texts}
    \label{fig:BibleScripts}
\end{figure}

\newpage
Figure~\ref{fig:Proverb}, ~\ref{fig:Ecc},  ~\ref{fig:Eccle} and ~\ref{fig:wisdomchp} shows the euclidean distance of chapters within the Bible texts which helps to find most similar chapters within the same book.

Amongst all scripts, chapters within Upanishads were most similar to themselves which is shown in  figure \ref{fig:Upanishads}. The diagonals represent the distance of a chapter to itself thus resulting in minimum distance of 0.

\begin{figure}
    \centering
    \begin{subfigure}[b]{0.49\textwidth}
        \includegraphics[width=\textwidth]{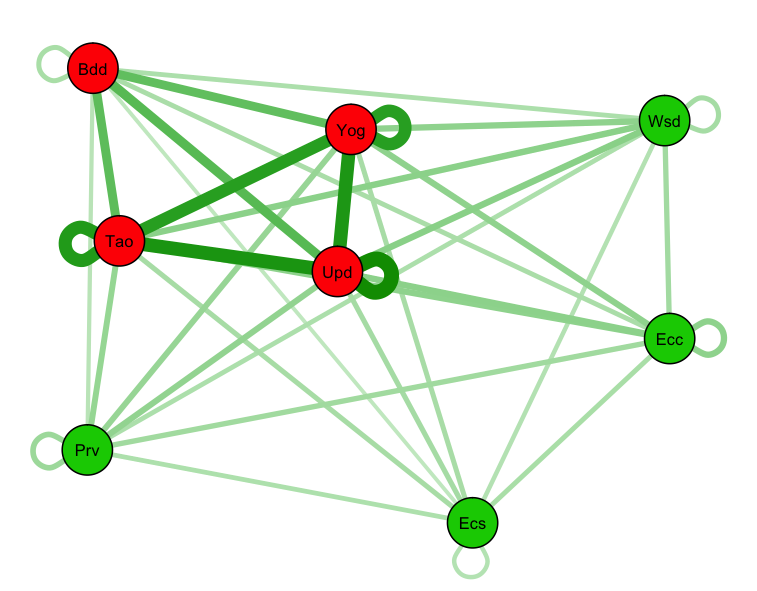}
        \caption{Graph network representation}
    \end{subfigure}
    \begin{subfigure}[b]{0.49\textwidth}
        \includegraphics[width=\textwidth]{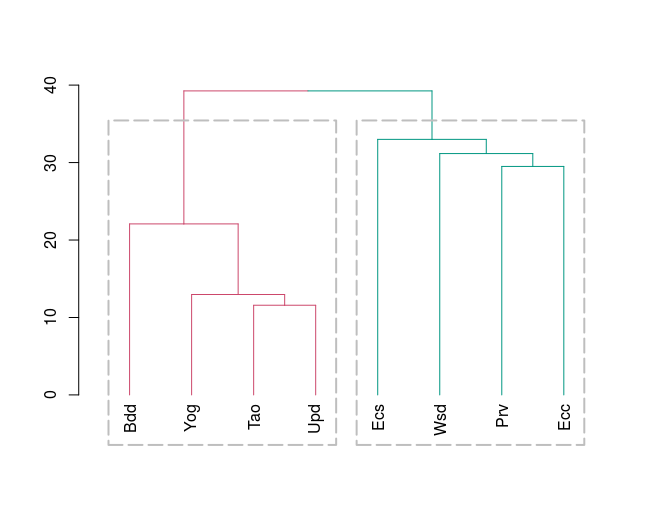}
        \caption{Tree Structure}
    \end{subfigure}
    \caption{Clustering with k = 2}
    \label{fig:multiscale2}
\end{figure}

\begin{figure}
    \centering
    \begin{subfigure}[b]{0.49\textwidth}
        \includegraphics[width=\textwidth]{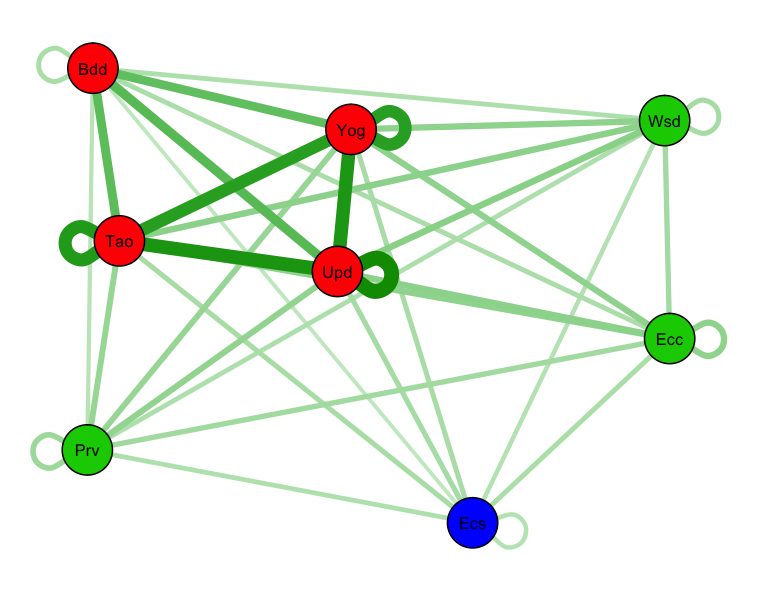}
        \caption{Graph network representation}
    \end{subfigure}
    \begin{subfigure}[b]{0.49\textwidth}
        \includegraphics[width=\textwidth]{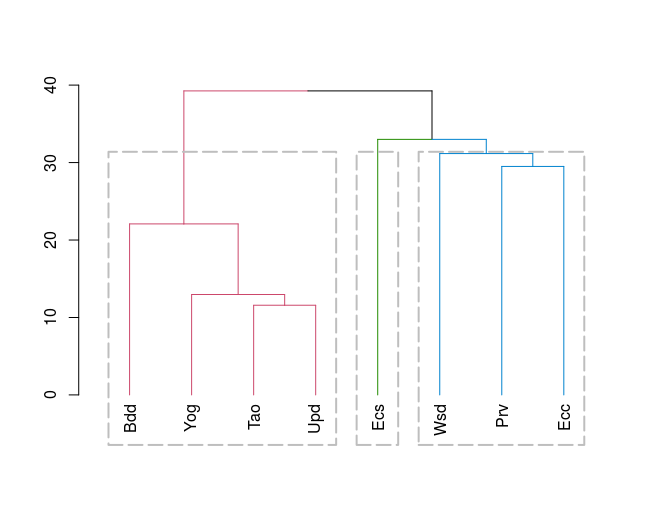}
        \caption{Tree Structure}
    \end{subfigure}
    \caption{Clustering with k = 3}
    \label{fig:multiscale3}
\end{figure}

\begin{figure}
    \centering
    \begin{subfigure}[b]{0.49\textwidth}
        \includegraphics[width=\textwidth]{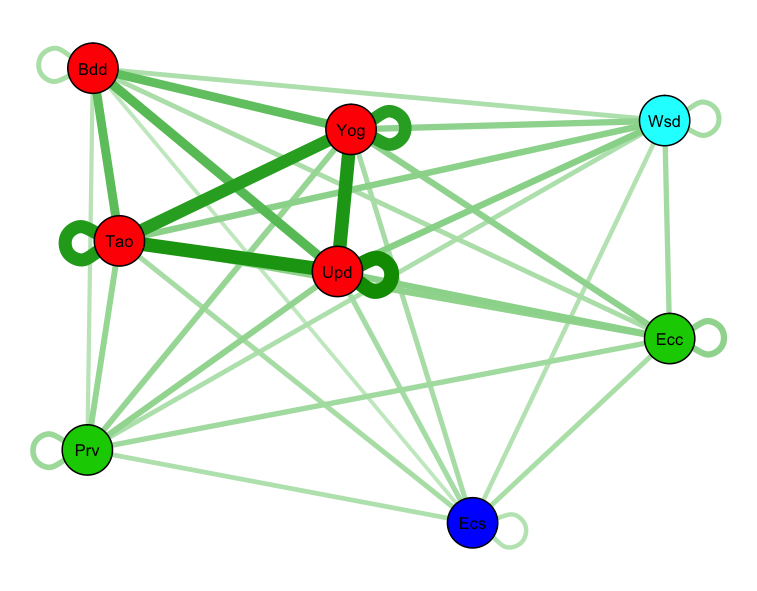}
        \caption{Graph network representation}
    \end{subfigure}
    \begin{subfigure}[b]{0.49\textwidth}
        \includegraphics[width=\textwidth]{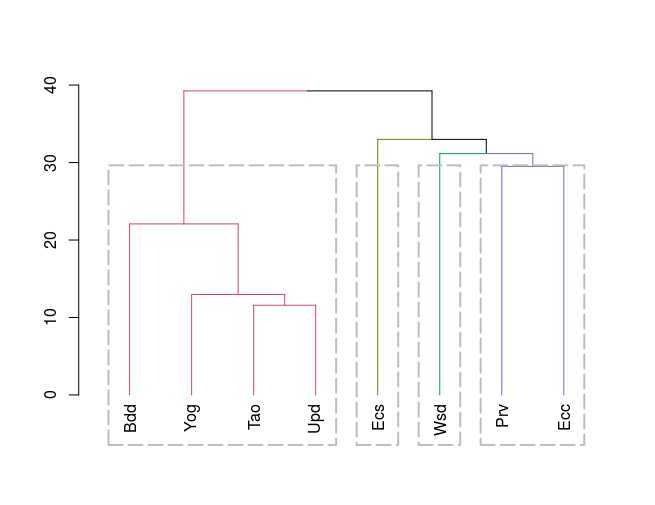}
        \caption{Tree Structure}
    \end{subfigure}
    \caption{Clustering with k = 4}
    \label{fig:multiscale4}
\end{figure}

\begin{figure}
    \centering
    \begin{subfigure}[b]{0.49\textwidth}
        \includegraphics[width=\textwidth]{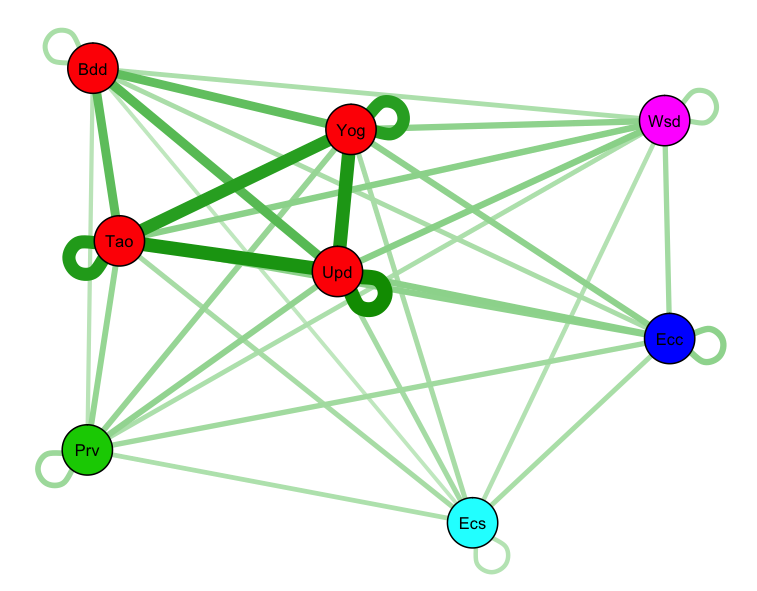}
        \caption{Graph network representation}
    \end{subfigure}
    \begin{subfigure}[b]{0.49\textwidth}
        \includegraphics[width=\textwidth]{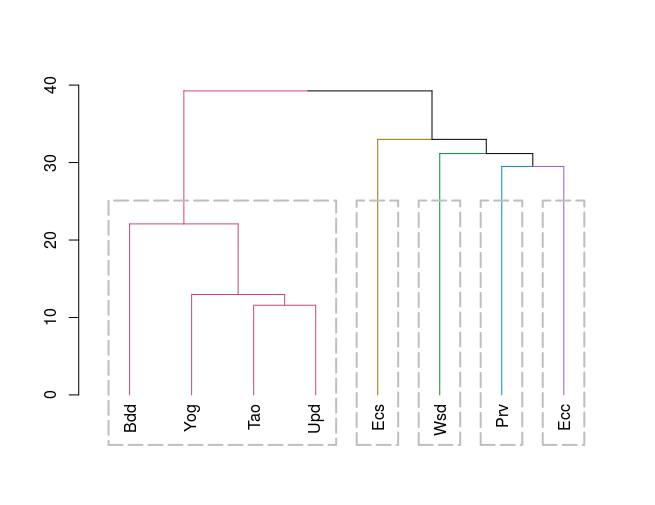}
        \caption{Tree Structure}
    \end{subfigure}
    \caption{Clustering with k = 5}
    \label{fig:multiscale5}
\end{figure}

\begin{figure}
    \centering
    \begin{subfigure}[b]{0.49\textwidth}
        \includegraphics[width=\textwidth]{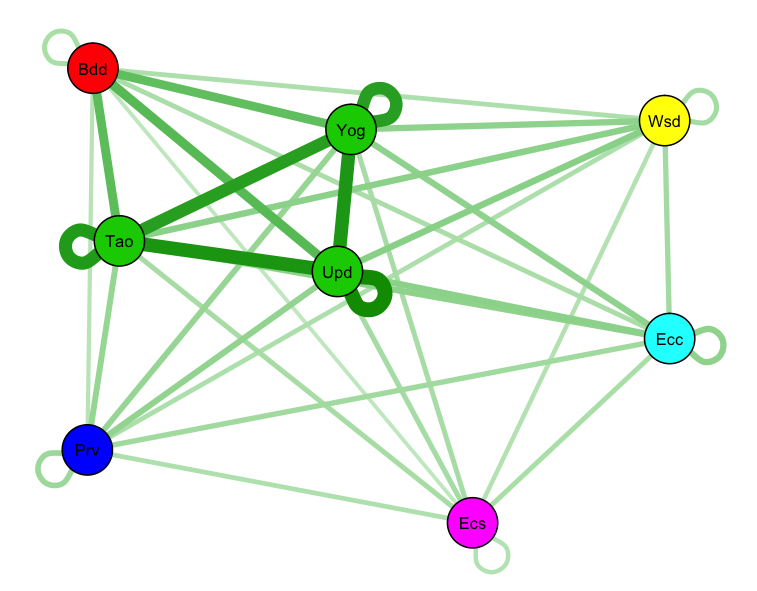}
        \caption{Graph network representation}
    \end{subfigure}
    \begin{subfigure}[b]{0.49\textwidth}
        \includegraphics[width=\textwidth]{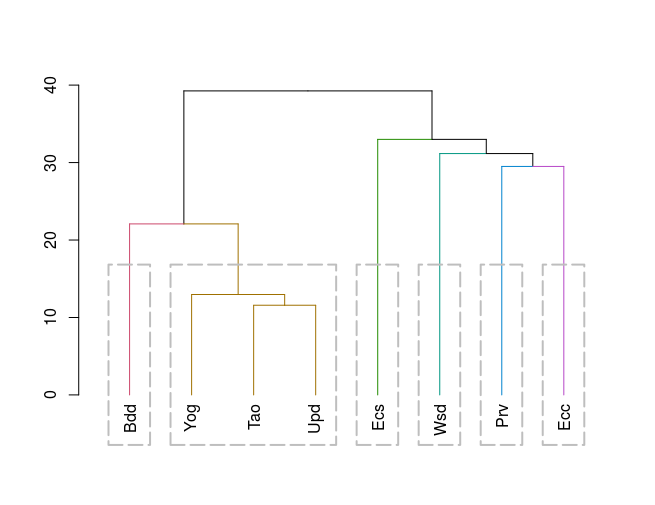}
        \caption{Tree Structure}
    \end{subfigure}
    \caption{Clustering with k = 6}
    \label{fig:multiscale6}
\end{figure}

\begin{figure}
    \centering
    \begin{subfigure}[b]{0.49\textwidth}
        \includegraphics[width=\textwidth]{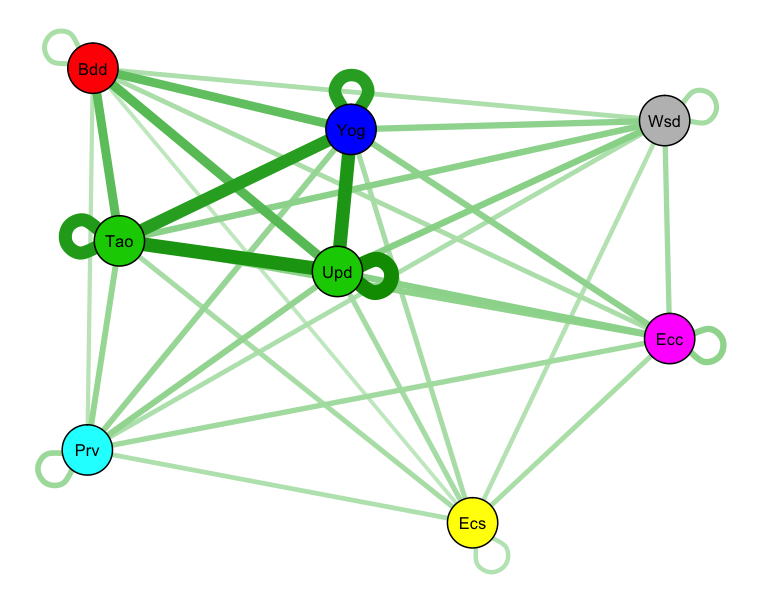}
        \caption{Graph network representation}
    \end{subfigure}
    \begin{subfigure}[b]{0.49\textwidth}
        \includegraphics[width=\textwidth]{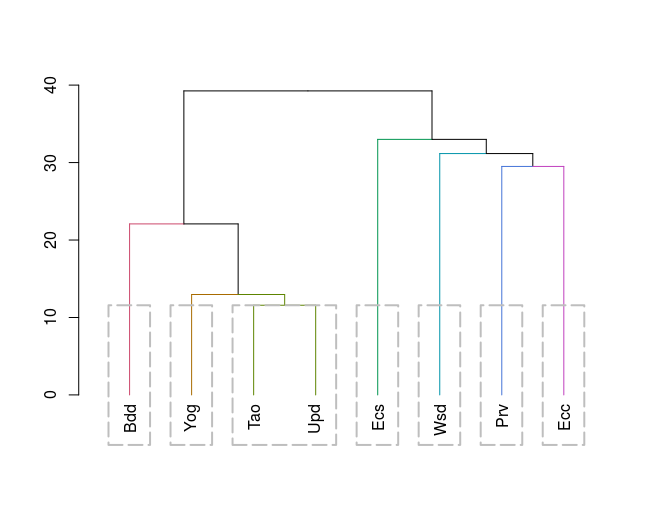}
        \caption{Tree Structure}
    \end{subfigure}
    \caption{Clustering with k = 7}
    \label{fig:multiscale7}
\end{figure}

The strength of similarity between different scripts can be found by visualizing k-means clustering results calculated from Euclidean distances in figure~\ref{fig:multiscale2}, ~\ref{fig:multiscale3} ~\ref{fig:multiscale4}, ~\ref{fig:multiscale5}, ~\ref{fig:multiscale6} and  ~\ref{fig:multiscale7}. Each node is the network graph represents a script and strength between two scripts is proportional to the width and brightness of edge. The cluster number(k) varies from two to seven and each figure represents groups of similarity for different k.
[Nodes : Bdd = Buddhism / Tao = TaoTeChing/  Upd = Upanishad/ Yoga = YogaSutra/ Prv = Proverb/ Ecc = Ecclesiastes/  Ecs = Ecclesiasticus/  Wsd = Wisdom]

\clearpage
\pagebreak

\clearpage
\pagebreak
The figure ~\ref{fig:multiscale3} represents that Asian texts are more similar to themselves as compared to the Biblical texts. As we increase the number of cluster from 2 to 7 we can visualize the similarities amongst Asian scripts. While moving from k=3 to 5 all biblical texts belong to different clusters which means even 4 biblical texts are quite different from each other. At the end Upanishads and Tao Te Ching are the most similar scripts as they belong to same cluster when k=7. 

 The performance of different supervised algorithms in predicting sacred text for any chapter in the scripture is shown in Table ~\ref{table:KNN}, ~\ref{table:SVM} and ~\ref{table:RF}.

\begin{table}[hbt]
\begin{center}
\begin{tabular}{|c|c|c|c|c|c|c|c|c|}
\hline
 & Buddhism & Ecclesiastes & Ecclesiasticus & Proverb & Tao & Upanishad & Wisdom & Yoga\\
 \hline
 Buddhism & \textbf{4} & 0 & 0 & 0 & 0 & 0 & 0 & 0 \\
 \hline
 Ecclesiastes & 0 & \textbf{0} & 0 & 0 & 0 & 0 & 0 & 0 \\
  \hline
 Ecclesiasticus & 0 & 0 & \textbf{0} & 0 & 0 & 0 & 1 & 0 \\
   \hline
 Proverb & 0 & 0 & 4 & \textbf{4} & 0 & 0 & 0 & 0 \\
    \hline
Tao & 0 & 0 & 0 & 0 & \textbf{0} & 0 & 0 & 0 \\
    \hline
Upanishad & 10 & 3 & 7 & 3 & 23 & \textbf{43} & 3 & 61 \\
    \hline
Wisdom & 0 & 0 & 1 & 0 & 0 & 0 & \textbf{1} & 0 \\
    \hline
Yoga & 1 & 0 & 0 & 0 & 0 & 0 & 0 & \textbf{5} \\
\hline
\end{tabular}
\caption{\label{table:KNN} Confusion matrix generated by KNN having accuracy = 0.339}
\end{center}
\end{table}

\begin{table}[hbt]
\begin{center}
\begin{tabular}{|c|c|c|c|c|c|c|c|c|}
\hline
 & Buddhism & Ecclesiastes & Ecclesiasticus & Proverb & Tao & Upanishad & Wisdom & Yoga\\
 \hline
 Buddhism & \textbf{1} & 0 & 0 & 0 & 0 & 0 & 0 & 0 \\
 \hline
 Ecclesiastes & 0 & \textbf{0} & 0 & 0 & 0 & 0 & 0 & 0 \\
  \hline
 Ecclesiasticus & 0 & 1 & \textbf{10} & 6 & 0 & 0 & 2 & 0 \\
   \hline
 Proverb & 0 & 0 & 0 & \textbf{0} & 0 & 0 & 0 & 0 \\
    \hline
Tao & 0 & 0 & 0 & 0 & \textbf{0} & 0 & 0 & 0 \\
    \hline
Upanishad & 0 & 0 & 1 & 1 & 0 & \textbf{0} & 0 & 0 \\
    \hline
Wisdom & 0 & 0 & 0 & 0 & 0 & 0 & \textbf{0} & 0 \\
    \hline
Yoga & 14 & 2 & 4 & 0 & 23 & 43 & 3 & \textbf{66} \\
\hline 
\end{tabular}
\caption{\label{table:SVM} Confusion matrix generated by SVM having accuracy = 0.435}
\end{center}
\end{table}

\begin{table}[hbt]
\begin{center}
\begin{tabular}{|c|c|c|c|c|c|c|c|c|}
\hline
 & Buddhism & Ecclesiastes & Ecclesiasticus & Proverb & Tao & Upanishad & Wisdom & Yoga\\
 \hline
 Buddhism & \textbf{8} & 0 & 0 & 0 & 0 & 0 & 0 & 0 \\
 \hline
 Ecclesiastes & 0 & \textbf{0} & 0 & 0 & 0 & 0 & 0 & 0 \\
  \hline
 Ecclesiasticus & 0 & 1 & \textbf{14} & 0 & 0 & 0 & 5 & 0 \\
   \hline
 Proverb & 0 & 0 & 1 & \textbf{7} & 0 & 0 & 0 & 0 \\
    \hline
Tao & 0 & 0 & 0 & 0 & \textbf{14} & 0 & 0 & 0 \\
    \hline
Upanishad & 7 & 0 & 0 & 0 & 8 & \textbf{43} & 0 & 8 \\
    \hline
Wisdom & 0 & 0 & 0 & 0 & 0 & 0 & \textbf{0} & 0 \\
    \hline
Yoga & 0 & 2 & 0 & 0 & 1 & 0 & 0 & \textbf{58} \\
\hline
\end{tabular}
\caption{\label{table:RF} Confusion matrix generated by Random Forest having accuracy = 0.8136}
\end{center}
\end{table}

Amongst all three supervised algorithms  Random Forest has highest accuracy of predicting which sacred text a fragment of spiritual literature comes from, as shown in  Table \ref{table:RF}. The Upnaishads and Yogasutra have the largest number of chapters in the corpus and random forest is accurately able to predict most of the chapters for these two sacred texts which SVM and KNN fail to identify.

\section{Conclusions}
After projecting Euclidean distances on various DTM (raw data DTM and normalized log DTM) we can conclude that the pattern of strong closeness exists among the different religious scripts. The similarity is driven by geography of origin of the religions. Bag of words is powerful to find the pattern of strong closeness between the four Asian religious scripts: Buddhism, Tao Te Ching, Upanishad and Yogasutra whose place of origin are geographical close. The two most similar scripts Tao Te Ching and Upanishad depicts the influence of two neighbouring countries China and India on their common religious teachings. 

An interesting potential work in this direction would be extracting main sematics features of the texts. Also, k-medoids using PAM can be implemented to observe the similarity between scripts. Using k-medoids ensures that the centers of clusters are actual points in the DTM and can give better results. This work also initiates the conversation about interesting results that be obtained from Markov models.

\bibliographystyle{acm}
\end{document}